%% file: main.tex
\documentclass[10pt,twocolumn,letterpaper]{article}

\usepackage{iccv}
\usepackage{times}
\usepackage{epsfig}
\usepackage{graphicx}
\usepackage{amsmath}
\usepackage{amssymb}

\usepackage{latexsym}

\usepackage{url}
\usepackage{booktabs}       
\usepackage{amsfonts}       
\usepackage{nicefrac}       
\usepackage{microtype}      
\usepackage{bbm}
\usepackage{color}
\usepackage{multirow}
\usepackage{caption}
\usepackage{subcaption}
\usepackage{enumitem}
\usepackage{hhline}
\usepackage{makecell}
\usepackage{pifont}

\usepackage[pagebackref=true,breaklinks=true,letterpaper=true,colorlinks,bookmarks=false]{hyperref}

\DeclareMathOperator*{\expect}{\mathbb{E}}
\newcommand\indicator[1]{\mathbbm{1}[#1]}
\newcommand{\friedaug}{\texttt{Fried-Augmented}}
\newcommand{\cmark}{\ding{51}}%
\newcommand{\xmark}{\ding{55}}%
\newcommand{\taskcma}{\textsc{cma}}
\newcommand{\tasknvs}{\textsc{nvs}}

\iffalse
\newcommand{\todo}[1]{}
\newcommand\eugeneie[1]{}
\newcommand\jason[1]{}
\newcommand\vihan[1]{}
\newcommand\rsents[1]{}
\newcommand\harsh[1]{}
\else
\newcommand{\todo}[1]{\textcolor{red}{TODO: {#1}}}
\newcommand\eugeneie[1]{[\textcolor{blue}{EI: {#1}}]}
\newcommand\jason[1]{[\textcolor{red}{JB: {#1}}]}
\newcommand\vihan[1]{[\textcolor{blue}{VJ: {#1}}]}
\newcommand\rsents[1]{[\textcolor{green}{HH: {#1}}]}
\newcommand\harsh[1]{[\textcolor{red}{HM: {#1}}]}

\fi

\iccvfinalcopy 

\def\iccvPaperID{6780} 

\ificcvfinal\fi
\begin{document}

\title{Transferable Representation Learning in Vision-and-Language Navigation}

\author{Haoshuo Huang\thanks{\ Authors contributed equally.} \qquad Vihan Jain \footnotemark[\value{footnote}] \qquad Harsh Mehta \qquad Alexander Ku \qquad Gabriel Magalhaes \\ Jason Baldridge \qquad Eugene Ie\\
Google Research\\
1600 Amphitheatre Parkway, Mountain View, CA 94043, United States\\
{\tt\small \{haoshuo, vihan, harshm, alexku, gamaga, jridge, eugeneie\}@google.com}
}

\maketitle
\ificcvfinal\thispagestyle{empty}\fi

\begin{abstract}
Vision-and-Language Navigation (VLN) tasks such as Room-to-Room (R2R) require machine agents to interpret natural language instructions and learn to act in visually realistic environments to achieve navigation goals. The overall task requires competence in several perception problems: successful agents combine spatio-temporal, vision and language understanding to produce appropriate action sequences. Our approach adapts pre-trained vision and language representations to relevant in-domain tasks making them more effective for VLN. Specifically, the representations are adapted to solve both a cross-modal sequence alignment and sequence coherence task. In the sequence alignment task, the model determines whether an instruction corresponds to a sequence of visual frames. In the sequence coherence task, the model determines whether the perceptual sequences are predictive sequentially in the instruction-conditioned latent space. By transferring the domain-adapted representations, we improve competitive agents in R2R as measured by the success rate weighted by path length (SPL) metric.
\end{abstract}

\input{introduction}

\input{related}

\input{data}

\input{negative_paths}

\input{representation_learning}

\input{navigation_agent}

\input{results}

\input{conclusion}

\section*{Acknowledgements}
\noindent
We thank the ICCV 2019 reviewers for their helpful reviews.

\input{main.bbl}
\end{document}

%% file: introduction.tex
\begin{figure}
  \centering   
  \includegraphics[clip, trim=3.3cm 3.1cm 4.8cm 1.5cm, width=\linewidth]{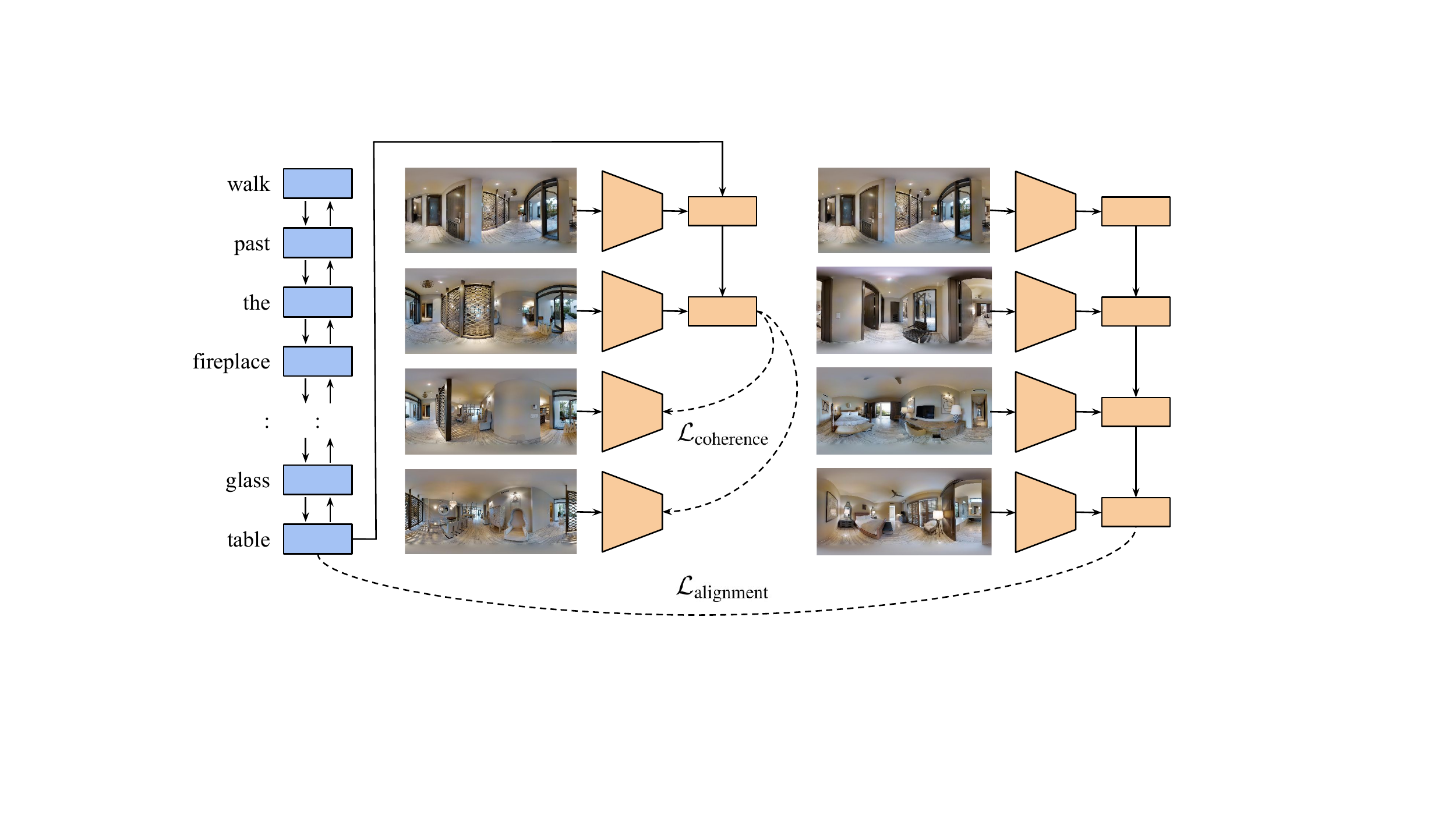}

\caption{ 
    To overcome the scarcity of high-quality human-annotated data, we propose auxiliary tasks, {\taskcma} and {\tasknvs}, that can be created by simple and effective negative mining. The representations learned by a model trained on both the tasks simultaneously, with a combined loss $\alpha \mathcal{L}_\text{alignment} + (1 - \alpha) \mathcal{L}_\text{coherence}$, are transferred over to agents learning the VLN navigation task. The RCM agent \cite{Wang:2018:RCM} so trained outperforms the existing published state-of-the-art agents.
    }
    \label{fig:intro_summary}
\end{figure}

\section{Introduction}
\label{sec:introduction}

Vision-and-Language Navigation (VLN) requires computational agents to represent and integrate both modalities and take appropriate actions based on their content, their alignment and the agent's position in the environment. VLN datasets have graduated from simple virtual environments \cite{Macmahon06walkthe} to photo-realistic environments, both indoors \cite{Anderson:2018:VLN} and outdoors
\cite{deVries:2018:TalkWalk,Cirik:2018:StreetView,Hermann:2019:StreeLang}. To succeed, VLN agents must internalize the (possibly noisy) natural language instruction, plan action sequences, and move in environments that dynamically change what is presented in their visual fields. These challenging settings bring simulation-based VLN work closer to real-world, language-based interaction with robots \cite{matuszek:2018}.




Along with these challenges come opportunities: for example, pre-trained linguistic and visual representations can be injected into agents before training them on example instructions-path pairs. Work on the Room-to-Room (R2R) dataset \cite{Anderson:2018:VLN} typically uses GloVe word embeddings \cite{Pennington:2014:GloVe} and features from deep image networks like ResNet \cite{He:2016:Resnet} trained on ImageNet \cite{Russakovsky:2015:ImageNet}. Associations between the input modalities are based on co-attention, with text and visual representations conditioned on each other. Since a trajectory spans multiple time steps, the visual context is often modeled using recurrent techniques like LSTMs \cite{Hochreiter:1997:LSTM} that combine features from the current visual field with historical visual signals and agent actions. The fusion of both modalities constitutes the agent's belief state. The agent relies on this belief state to decide which action to take, often relying on reinforcement learning techniques like policy gradient \cite{Williams:1992:PolicyGradient}.

Unfortunately, due to domain shift, the pre-trained models are poor matches for R2R's instructions and visual observations. Furthermore, human-annotated data is expensive to collect and there are relatively few instruction-path pairs (e.g. R2R has just 7,189 paths with instructions). This greatly reduces the expected benefit of fine-tuning \cite{Girshick:2014:Finetuning,Zeiler:2014:Pretraining} on the navigation task itself. Our contribution is to define auxiliary, discriminative learning tasks that exploit the environment \textit{before} agent training. Our high-quality augmentation strategy adapts the out-of-domain pre-trained representations and allows the agent to focus on learning how to act rather than struggling to bridge representations \textit{while} learning how to act. It furthermore allows us to rank and better exploit the outputs of generative strategies used previously \cite{Fried:2018:Speaker}.



We present three main contributions. First, we define two in-domain auxiliary tasks: Cross-Modal Alignment ({\taskcma}), which involves assessing the fit between a given instruction-path pair, and Next Visual Scene ({\tasknvs}), which involves predicting latent representations of future visual inputs in the path. Neither task requires additional human annotated data as they are both trained with cheap negative mining techniques following Huang \etal \cite{Huang:2019:Discriminator}. Secondly, we propose methods to train models on the two tasks: \emph{alignment-based} similarity scores for {\taskcma} and contrastive predictive coding \cite{Oord2018RepresentationLW} for {\tasknvs}. A model trained on {\taskcma} and {\tasknvs} is not only able to learn cross-modal alignments, but is also able to correctly differentiate between high-quality and low-quality instruction-path pairs in the augmented data introduced by Fried \etal \cite{Fried:2018:Speaker}. Finally, we show that representations learned by this model can be transferred to two competitive navigation agents, Speaker-Follower \cite{Fried:2018:Speaker} and Reinforced Cross-Modal \cite{Wang:2018:RCM}, to outperform their previously established results. We also found that our domain-adapted agent outperforms the known state-of-the-art agent at the time by 5\% absolute measure in SPL.

%% file: related.tex
\section{Related Work}

\noindent
\textbf{Vision-and-Language Grounding} There is much prior work in the intersection of computer vision and natural language processing \cite{2015:Xu:ShowAttendTell,Karpathy:2015:VisualAlignmentCaptions,Mao:2016:GenerationComprehensionObjects,Hu:2016:SegmentationNLE}. A highly related class of tasks centers around generating captions for images and videos \cite{Donahue:2017:LTRCNCaption,Fang:2015:VisualCaptions,Vinyals:2015:ShowAndTell,Wang:2018:VideoCV,Yu:2016:VideoParagraphCaptioning}. In Visual Question Answering \cite{Antol:2015:VQA,Yang:2016:StackedANQnA} and Visual Dialog \cite{das:2017:VisualDialog}, models generate single-turn and multi-turn responses by co-grounding vision and language. In contrast to these tasks, VLN agents are embodied in the environment and must combine language, scene, and spatio-temporal understanding.

\noindent
\textbf{Embodied Agent Navigation} Navigation in realistic 3D environments has also received increased interest recently \cite{Toshev:2018:SemanticNavigation,Hemachandra:2015:VLN,Mirowski:2018:StreetLearn,Zhu:2017:IndoorNavigation}. Advances in vision-and-language navigation have accelerated with the introduction of the Room-to-Room (R2R) dataset and associated attention-based sequence-to-sequence baseline \cite{Anderson:2018:VLN}. Fried \etal \cite{Fried:2018:Speaker} used generative approaches to augment the instruction-path pairs and proposed a modified beam search for VLN. Wang \etal \cite{Wang:2018:RCM} introduced innovations around multi-reward RL with imitation learning and co-grounding in the visual and text modality. While the two approaches reused pre-trained vision and language modules directly in the navigation agent, our contribution shows that these pre-trained components can be further enhanced by adapting them to related auxiliary tasks before employing them in a VLN agent.

%% file: data.tex
\section{The Room-to-Room Dataset}
\label{sec:data}

The Room-to-Room (R2R) dataset \cite{Anderson:2018:VLN} is based on 90 houses from the Matterport3D environments \cite{Matterport3D} each defined by an undirected graph. The nodes are locations where egocentric photo-realistic panoramic images are captured and the edges define the connections between locations. The dataset consists of language instructions paired with reference paths, where each path is a sequence of graph nodes. Each path is associated with 3 natural language instructions collected using Amazon Mechanical Turk with an average token length of 29 from a dictionary of 3.1k unique words. Paths collected are longer than 5m and contain 4 to 6 edges. The dataset is split into a training set, two validation sets and a test set. One validation set includes new instructions on environments overlapping with the training set (Validation Seen), and the other is entirely disjoint from the training set (Validation Unseen). Evaluation on the validation unseen set and the test set assess the agent's full generalization ability. Metrics for assessing agents performance include:

\begin{itemize}
    \setlength \itemsep{0em}
    \item \textit{Path Length (PL)} measures the total length of the predicted path. (The reference path's length is optimal.)
    \item \textit{Navigation Error (NE)} measures the distance between the last nodes in the predicted and the reference paths.
    \item \textit{Success Rate (SR)} measures how often the last node in the predicted path is within some threshold distance $d_{th}$ of the last node in the reference path.
    \item \textit{Success weighted by Path Length (SPL)} \cite{Anderson:2018:Evaluation} measures whether the SR success criteria was met, weighted by the normalized path length.
\end{itemize}

\noindent
SPL is the best metric for ranking agents as it takes into account the path taken, not just whether goal was reached \cite{Anderson:2018:Evaluation}. This is evident with (invalid) entries on the R2R leaderboard that use beam search often achieving high SR but low SPL because they wander all around before stopping.

%% file: negative_paths.tex
\section{Mining Negative Paths}
\label{sec:negative_paths}

VLN tasks are composed of instruction-path pairs, where a path is a sequence of connected locations along with their corresponding perceptual contexts. The core task is to train agents to follow the provided instructions. However, auxiliary tasks could help adapt out-of-domain language and vision representations to be relevant to the navigation domain. We follow two principles in designing these auxiliary tasks: they should not involve any additional human annotations and they should use and update representations needed for downstream navigation tasks.

The crux of our auxiliary tasks is the observation that the given human generated instructions are specific to the paths described. Given the diversity and relative uniqueness of the properties of different rooms and the trajectories of different paths, it is highly unlikely that the original instruction will correspond well to automatically mined negative paths. As such, given a visual path and a high quality human generated instruction, it is easy to create various incorrect paths by random path sampling or random walks from start or end nodes, to name a few. For a given instruction-path pair, we sample negatives by keeping the same instruction but altering the path sequence in one of three ways.

\begin{itemize}
    \setlength \itemsep{0em}
    \item \textit{Path Substitution (PS)}: randomly pick other paths from the same environment as negatives.
    \item \textit{Random Walks (RW)}: sample random paths of the same length as the original path that either (1) start at the same location and end sufficiently far from the original path or (2) end at the same location and start sufficiently far from the original path. We use a threshold of 5 meters to make sure the path has significant difference.  
    \item \textit{Partial Reordering (PR)}: keep the first and last nodes in the path fixed and randomly shuffle the rest.
\end{itemize}

\noindent
These three strategies create increasingly more challenging negative examples. PS pairs have only incidental connection between the text and the perceptual sequence, RW pairs share one or the other end point, and PR pairs have the same perceptual elements in a new (and incoherent) order.

%% file: representation_learning.tex
\section{Representation Learning}
\label{sec:representation_learning}


Using the mined negative paths, we train models for two auxiliary tasks that exploit the data in complementary ways. The first is a two-tower model \cite{Gillick:2018:DualEncoders,Serban:2018:DualEncoders} with a cross-modal alignment module. This model produces similarity scores that reflect the semantic similarity between visual and language sequences. The second is a model that optimizes pairwise sequence coherence by predicting latent representations of future visual scenes, conditioned on the language sequence and a partial visual sequence. We furthermore train these models on both tasks with a combined loss. This fine tunes the representations to domain-specific language and interior environments relevant to the R2R dataset, and associates language to the visual scenes the agent will experience during the full navigation problem. 


\subsection{Task 1: Cross-Modal Alignment (\taskcma)}

An agent's ability to navigate a visual environment using language instructions is closely associated with its capacity to align semantically similar concepts across the two modalities. Given an instruction like ``\textit{Turn right and move forward around the bed, enter the bathroom and wait there.}'', the agent should match the word \textit{bed} with a location on the path that has a bed in the agent's egocentric view; doing so will help orient the agent and allow it to better follow further instructions. To this end, we create a cross-modal alignment task (denoted as {\taskcma}) that involves discriminating positive instruction-path pairs from negative pairs. The discriminative model is based on an \textit{alignment-based} similarity score that encourages the model to map perceptual and textual signals in two sequences.

\subsection{Task 2: Next Visual Scene (\tasknvs)}

Research in sensory and motor processing suggests that the human brain predicts (anticipates) future states in order to assist decision making \cite{Luca2016Optimal,Bubic2010Prediction}. Similarly, agents can benefit if they learn to predict expected future states given the current context at a given point in the course of navigation.
While it is challenging to predict high-dimensional future states, methods like Contrastive Predictive Coding (CPC) \cite{Oord2018RepresentationLW} circumvent this by working in lower dimensional latent spaces.
With CPC, we add a probabilistic contrastive loss to our adaptation model. This induces a latent space that captures visual information useful for predicting future visual observations, enabling the visual network to adapt to the R2R environment. In the \tasknvs\ task, the model's current state is used to predict the latent space representation of future $k$ steps (in this work, we use $k={1,2}$).
The negatives from {\taskcma} are used as negatives to compute the InfoNCE \cite{Oord2018RepresentationLW} loss during training (see next section for details).

\subsection{Model Architecture}
\label{subsec:model_arch}
For consistency with the navigation agent model (Sec.~\ref{sec:navigation_agent}), we use a two-tower architecture to encode the two sequences, with one tower encoding the token sequence in the instruction and the other tower encoding the visual sequence.

\textbf{Language Encoder.} 
Instructions $\mathcal{X}={x_1, x_2, ..., x_n}$ are initialized with pre-trained GloVe word embeddings \cite{Pennington:2014:GloVe}. These embeddings are fine-tuned to solve the auxiliary tasks and transferred to the agent to be further fine-tuned to solve the VLN challenge. We restrict the GloVe vocabulary to tokens that occur at least five times in the training instructions. All out-of-vocabulary tokens are mapped to a single out-of-vocabulary identifier. The token sequence is encoded using a bi-directional LSTM \cite{Schuster1997BidirectionalRN} to create $H^X$ following:

\begin{small}
\begin{align}
    H^X &= [h_1^X; h_2^X; ...; h_n^X]  \\
    h_t^X &= \sigma(\overrightarrow{h}_t^X, \overleftarrow{h}_t^X)  \\
    \overrightarrow{h}_t^X &= LSTM(x_t, \overrightarrow{h}_{t-1}^X)  \\
    \overleftarrow{h}_t^X &= LSTM(x_t, \overleftarrow{h}_{t+1}^X)
\end{align}
\end{small}

\noindent
where the $\sigma$ function is used to combine the output of forward and backward LSTM layers.

\textbf{Visual Encoder.}
As in Fried \etal \cite{Fried:2018:Speaker}, at each time step $t$, the agent perceives a 360-degree panoramic view at its current location. The view is discretized into $k$ view angles ($k=36$ in our implementation, 3 elevations by 12 headings at 30-degree intervals).
The image at view angle $i$, heading angle $\phi$ and elevation angle $\theta$ is represented by a concatenation of the pre-trained CNN image features with the 4-dimensional orientation feature [sin $\phi$; cos $\phi$; sin $\theta$; cos $\theta$] to form $v_{t,i}$.
The visual input sequence $\mathcal{V}={v_1, v_2, ..., v_m}$ is encoded using a LSTM to create $H^V$ following:

\begin{small}
\begin{align}
    H^V &= [h_1^V; h_2^V; ...; h_m^V]  \\
    h_t^V &= LSTM(v_t, h_{t-1}^V)
\end{align}
\end{small}

\noindent
where $v_t=\text{Attention}(h_{t-1}^V, v_{t,1..k})$ is the attention-pooled representation of all view angles using previous agent state $h_{t-1}$ as the query.

\textbf{Training Loss.}
For {\taskcma}, the alignment-based similarity score is computed as follows:

\begin{small}
\begin{align}
    A &= H^X({H^V})^T  \label{eq:A_compute} \\
    \{c\}_{l=1}^{l=X} &= \text{softmax}(A^l) \cdot A^l  \label{eq:A_softmax} \\
    \text{score} &= \text{softmin}(\{c\}_{l=1}^{l=X}) \cdot \{c\}_{l=1}^{l=X}  \label{eq:A_softmin}
\end{align}
\end{small}

\noindent
where $(.)^T$ is matrix transpose transformation, $A$ is the alignment matrix whose dimensions are $[n, m]$ and $A^l$ is the $l$-th row vector in $A$. Eq.~\ref{eq:A_softmax} corresponds to taking a softmax along the columns and summing the columns. This amounts to column-wise content-based pooling. Then we apply the softmin operation along the rows and sum the rows up to obtain a scalar in Eq.~\ref{eq:A_softmin}. Intuitively, maximizing this score for positive instruction-path pairs encourages the learning algorithm to construct the best worst-case sequence alignment between the two sequences in the latent space. The training objective for {\taskcma} is to minimize the cross entropy loss $\mathcal{L}_\text{alignment}$.

The InfoNCE \cite{Oord2018RepresentationLW} loss for {\tasknvs} is computed as follows:

\begin{small}
\begin{align}
    \mathcal{L}_\text{coherence} &= -\expect_F\left[\log \dfrac{f(v_{t+k}, h_t^V)}{\sum_{v_j \in F} f(v_j, h_t^V)}\right]  \label{eq:coherence_loss} \\
    f(v_{t+k}, h_t^V) &= \exp({v_{t+k}}^T W_k h_t^V)
\end{align}
\end{small}

\noindent
where $v_{t+k}$ is the latent representation of visual input at time step $t+k$, $h_t^V$ is the visual-encoder LSTM's output at time step $t$ which summarizes all $v_{\leq t}$, $W_k$ are learnable parameters which are different for different values of $k$ (we choose $k={1,2}$ in our experiments). For a given $h_t^V$, there is exactly one positive sample in the set $F$, the negative samples can be chosen from negative instruction-path pairs as mined in Sec.~\ref{sec:negative_paths}. The loss in Eq.~\ref{eq:coherence_loss} is the categorical cross-entropy of classifying the positive sample correctly.

Finally, the model is trained to minimize the combined loss $\alpha \mathcal{L}_\text{alignment} + (1 - \alpha) \mathcal{L}_\text{coherence}$.

%% file: navigation_agent.tex
\section{Navigation Agent}
\label{sec:navigation_agent}

For comparisons with established models, we reimplemented the Speaker Follower agent of Fried \etal \cite{Fried:2018:Speaker} (denoted as SF agent from hereon) and Reinforced Cross-Modal Matching agent of Wang \etal \cite{Wang:2018:RCM} (denoted as RCM agent from hereon) for our experiments.

\subsection{Navigator}
\label{sec:navigator}

The navigator learns a policy $\pi_{\theta}$ over parameters $\theta$ that map the natural language instruction $\mathcal{X}$ and the initial visual scene $v_1$ to a sequence of actions $a_{1..T}$. The language and visual encoder of the navigator are the same as described in Sec.~\ref{subsec:model_arch}. The actions available to the agent at time $t$ are denoted as $u_{t,1..l}$, where $u_{t,j}$ is the representation of the navigable direction $j$ from the current location obtained similarly to $v_{t,i}$ \cite{Fried:2018:Speaker}. The number of available actions, $l$, varies per location, since graph node connectivity varies. As in \cite{Wang:2018:RCM}, the model predicts the probability $p_d$ of each navigable direction $d$ using a bilinear dot product:

\begin{small}
\begin{align}
p_d &= \text{softmax}([h_t^V; c_t^{\text{text}}; c_t^{\text{visual}}]W_c(u_{t,d}W_u)^T)\\
c_t^{\text{text}} &= \text{Attention}(h_t^V, h^X_{1..n}) \\
c_t^{\text{visual}} &= \text{Attention}(c_t^{\text{text}}, v_{t, 1..k})
\end{align}
\end{small}

\subsection{Learning}
\label{sec:learning}

The SF agent is trained using \textit{student forcing} \cite{Fried:2018:Speaker} where actions are sampled from the model during training, and supervised using a shortest-path action to reach the goal.

For the RCM agent, learning is performed in two separate phases, (1) behavioral cloning \cite{Bain:1999:Cloning,Wang:2018:RCM,Daftry:2016:TransferablePolicies} and (2) REINFORCE policy gradient updates \cite{Williams:1992:PolicyGradient}. The agent's policy is initialized using behavior cloning to maximally use the available expert demonstrations. This phase constrains the learning algorithm to first model state-action spaces that are most relevant to the task, effectively warm starting the agent with a good initial policy. No reward shaping is required during this phase as behavior cloning corresponds to solving the following maximum-likelihood problem:

\begin{equation}
\max_{\theta} \sum_{(s,a) \in \mathcal{D}} \log \pi_{\theta}(a|s)
\end{equation}

\noindent
where $\mathcal{D}$ is the demonstration data set.

Once the model is initialized to a reasonable policy with behavioral cloning, we further update the model via standard policy gradient updates by sampling action sequences from the agent's behavior policy. As in standard policy gradient updates, the model minimizes the loss function $\mathcal{L}^{\text{PG}}$ whose gradient is the negative policy gradient estimator \cite{Williams:1992:PolicyGradient}:

\begin{equation}
\mathcal{L}^{\text{PG}} = -\hat{\expect}_t[\log \pi_{\theta}(a_{t}|s_{t}) \hat{A}_t]
\end{equation}

\noindent
where the expectation $\hat{\expect}_t$ is taken over a finite batch of sample trajectories generated by the agent's stochastic policy $\pi_{\theta}$. Furthermore, for variance reduction, we scale the gradient using the advantage function $\hat{A}_t = R_t - \hat{b}_t$ where $R_t=\sum_{i=t}^\infty \gamma^{i-t}r_i$ is the observed $\gamma$-discounted episodic return and $\hat{b}_t$ is the estimated value of agent's current state at time $t$. Similar to \cite{Wang:2018:RCM}, the immediate reward at time step $t$ in an episode of length $T$ is given by:

\begin{equation}
\label{eq:denseSRreward}
r(s_t, a_t) =
    \begin{cases}
        d(s_t, r_{|R|}) - d(s_{t+1}, r_{|R|}) & \text{if $t < T$} \\
        \indicator{d(s_T, r_{|R|}) \leq d_{th}} & \text{if $t = T$}
    \end{cases}
\end{equation}

\noindent
where $d(s_t, r_{|R|})$ is the distance between $s_t$ and target location $r_{|R|}$, $\indicator{\cdot}$ is the indicator function, $d_{th}$ is the maximum distance from $r_{|R|}$ that the agent is allowed to terminate for it to be considered successful.

The models are trained using mini-batch gradient descent. For RCM agent, our experiments show that interleaving behavioral cloning and policy gradient training phases improves performance on the validation set. Specifically we interleaved each policy gradient update batch with $K$ behaviour cloning batches, with the value of $K$ decaying exponentially, such that the training strategy asymptotically becomes only policy gradient updates.

%% file: results.tex
\section{Results}
\label{sec:results}

\begin{table}
\centering
\begin{tabular}{lll|r}
 PS        & PR        & RW         & AUC \\
\hline
 \cmark    &           &            & 64.5 \\
           & \cmark    &            & 60.5 \\
           &           & \cmark     & 63.1 \\
 \cmark    & \cmark    &            & 72.1 \\
           & \cmark    & \cmark     & 66.0 \\ 
 \cmark    &           & \cmark     & 70.8 \\
 \cmark    & \cmark    & \cmark     & 72.0 \\
\end{tabular}
\caption{Results on training in different combinations of datasets and evaluating against validation dataset containing PR and RW negatives only.\label{tab:sampling-strategy}}
\end{table}

\begin{table*}
\centering
\setlength\tabcolsep{4.5pt}
\begin{tabular}{llcccccccc}\\[-0.8em]
                              &  & \multicolumn{4}{c}{\textbf{Validation Seen}} & \multicolumn{4}{c}{\textbf{Validation Unseen}}\\\cmidrule(lr{0.1cm}){3-6}\cmidrule(l{0.1cm}){7-10}
Dataset size & Strategy                            & PL    & NE $\downarrow$   & SR $\uparrow$   & \underline{SPL} $\uparrow$  & PL    & NE $\downarrow$   & SR $\uparrow$   & \underline{SPL} $\uparrow$   \\[0.2em]\Xhline{2\arrayrulewidth}            
\multirow{2}{*}{1\%}   & Top                   & 11.1 & 8.5 & 21.2 & 17.6 & 11.2 & 8.5 & 20.4 & 16.6  \\
                       & Bottom                & 10.7 & 9.0 & 16.3 & 13.1 & 10.8 & 8.9 & 15.4 & 14.1  \\
\hline 
\multirow{2}{*}{2\%}   & Top                   & 11.7 & 7.9 & 25.5 & 21.0 & 11.3 & 8.2 & 22.3 & 18.5 \\
                       & Bottom                & 14.5 & 9.1 & 17.7 & 12.7 & 11.4 & 8.4 & 17.5 & 14.1  \\   
\end{tabular}
\caption{Results for Validation Seen and Validation Unseen, when trained with a small fraction of {\friedaug} ordered by scores given by model trained on {\taskcma}. SPL and SR are reported as percentages and NE and PL in meters.}
\label{tab:r2r-small-dataset}
\end{table*}

\subsection{Experimental Setup}
In our experiments, we use a 2-layer bi-directional LSTM for the instruction encoder where the size of LSTM cells is 256 units in each direction. The inputs to the encoder are 300-dimensional embeddings initialized using GLoVe and fine-tuned during training. For the visual encoder, we use a 2-layer LSTM with a cell size of 512 units. The encoder inputs are image features derived as mentioned in Sec.~\ref{subsec:model_arch}. The cross-modal attention layer size is 128 units. To train the model on auxiliary tasks, we use Momentum optimizer with a learning rate of $10^{-2}$ that decays at a rate of 0.8 every 0.5 million steps. The SF navigation agent is trained using Momentum optimizer while RCM agent is trained using Adam optimizer with learning rate decaying at a rate of 0.5 every 0.2 million steps. We use a learning rate of $10^{-5}$ during agent training if the agent is warm-started with pre-trained components of the model trained on auxiliary tasks, otherwise we use learning rate of $10^{-4}$.

\begin{figure*}
\centering
\includegraphics[clip, trim=3cm 5cm 3.5cm 4cm, width=0.75\textwidth]{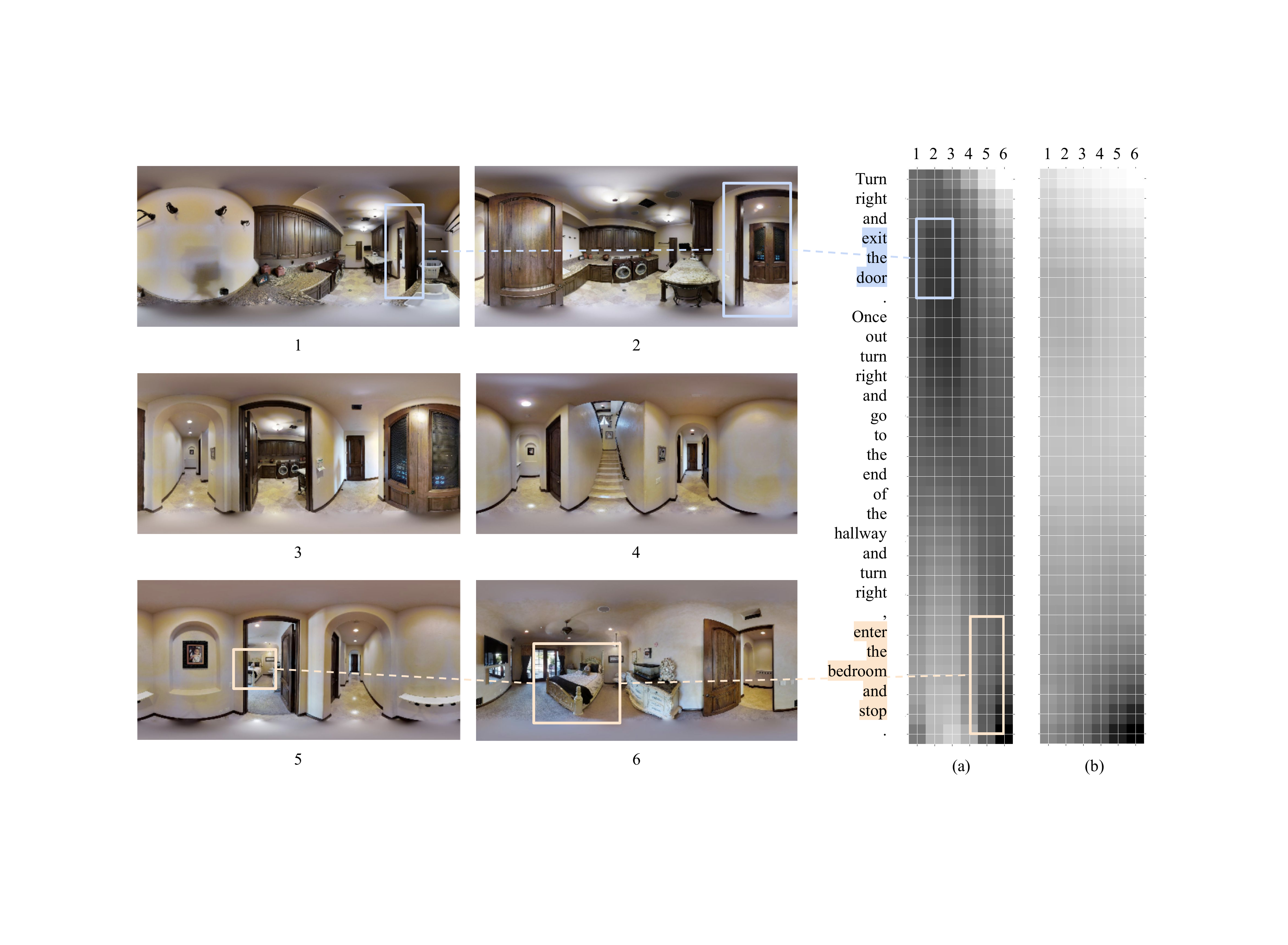}
\caption{Alignment matrix (Eq.~\ref{eq:A_compute}) for model trained on the dataset containing (a) PS, PR, RW negatives (b) PS negatives only. Note that darker means higher alignment. \label{fig:curriculum-learning-attention}}
\end{figure*}

\subsection{Training on Auxiliary Tasks}
Recently, Fried \etal \cite{Fried:2018:Speaker} introduced an augmented dataset (referred to as {\friedaug} from now on) that is generated by using a speaker model and they show that the models trained with both the original data and the machine-generated augmented data improves agent success rates. On manual inspection, we found that while many paths in {\friedaug} have clear starting \textit{or} ending descriptions, the middle segments of the instructions are often noisy and have little connection to the path they are meant to describe. Here we show that our model trained on {\taskcma} is able to differentiate between high-quality and low-quality instruction-path pairs in {\friedaug}.

In line with the original R2R dataset \cite{Anderson:2018:VLN}, we create three splits for each of the negative sampling strategies defined in Section \ref{sec:representation_learning} -- a training set from paths in R2R \textit{train} split, a validation seen set from paths in R2R \textit{validation seen} and a validation unseen set from paths in R2R \textit{validation unseen} split. The paths in the original R2R dataset are used as positives and there are 10 negatives for each positive with 4 of those negatives sampled using PS and 3 each using RW and PR respectively. A model trained on the task {\taskcma} learns to differentiate aligned instruction-path pairs from the misaligned pairs. We also studied the three negative sampling strategies summarized in Table~\ref{tab:sampling-strategy}.

\textbf{Scoring generated instructions.} We use this trained model to rank all the paths in {\friedaug} and train the RCM agent on different portions of the data. Table \ref{tab:r2r-small-dataset} gives the performance when using the best 1\% versus the worst 1\%, and likewise for the best and worst 2\%. Using high-quality examples--as judged by the model--outperforms the ones trained using low-quality examples. Note that the performance is low in both cases because none of the original human-created instructions were used---what is important is the relative performance between examples judged higher or lower. This clearly indicates that the model scores instruction-path pairs effectively.

\textbf{Visualizing Cross-Modal Alignment.} Fig.~\ref{fig:curriculum-learning-attention} gives the alignment matrix $A$ (Eq.~\ref{eq:A_compute}) from the model trained on {\taskcma} for a given instruction-path pair to try to better understand how well the model learns to align the two modalities as hypothesized. As a comparison point, we also plot the alignment matrix for a model trained on the dataset with PS negatives only. While scoring PR and RW negatives may require carefully aligning the full sequence in the pair, it is easier to score PS negatives by just attending to first or last locations on the path. It is expected that the model trained on the dataset containing only PS negatives will exploit these easy-to-find patterns in negatives and make predictions without carefully attending to full instruction-path sequence.

The figure shows the difference between cross-modal alignment for the two models. While there is no clear alignment between the two sequences for the model trained with PS negatives only (except maybe towards the end of sequences, as expected), there is a visible diagonal pattern in the alignment for the model trained on all negatives in {\taskcma}. In fact, there is appreciable alignment at the correct positions in the two sequences, \eg, the phrase \textit{exit the door} aligns with the image(s) in the path containing the object \textit{door}, and similarly for the phrase \textit{enter the bedroom}.

\textbf{Improvements from Adding Coherence Loss.}
Finally we show that training a model on {\taskcma} and {\tasknvs} simultaneously improves the model's performance when evaluated on {\taskcma} alone. The model is trained using combined loss $\alpha \mathcal{L}_\text{alignment} + (1 - \alpha) \mathcal{L}_\text{coherence}$ with $\alpha=0.5$ and is evaluated on its ability to differentiate incorrect instruction-path pairs from correct ones. As noted earlier, PS negatives are easier to discriminate, therefore, to keep the task challenging, the validation sets were limited to contain validation splits from PR and RW negative sampling strategies only. The area-under ROC curve (AUC) is used as the evaluation metric. The results in Table \ref{tab:task1_task2} demonstrate that adding $\mathcal{L}_\text{coherence}$ as auxiliary loss improves the model's performance on {\taskcma} by 7\% absolute measure.


\begin{table}
    \centering
    \begin{tabular}{lrr}\\[-0.9em]
      Training   & Val. Seen & Val. Unseen \\\hline
      {\taskcma} & 82.6 &  72.0 \\
      {\tasknvs} & 63.0 & 62.1 \\
      {\taskcma} + {\tasknvs} & \textbf{84.0} & \textbf{79.2} \\ 
    \end{tabular}
    \caption{
    AUC performance when the model is trained on different combinations of the two tasks and evaluated on the dataset containing only PR and RW negatives. 
    }
    \label{tab:task1_task2}
\end{table}

\subsection{Transfer Learning to Navigation Agent}

The language and visual encoders in the RCM navigation agent (Sec.~\ref{sec:navigation_agent}) are warm-started from the model trained on {\taskcma} and {\tasknvs} simultaneously. The agent is then allowed to train on R2R \textit{train} and {\friedaug} as other existing baseline models do. We call this agent \textit{ALTR} -- to mean an \textbf{A}gent initialized by \textbf{L}earned \textbf{T}ransferable \textbf{R}epresentations from auxiliary tasks. The standard testing scenario of the VLN task is to train the agent in seen environments and then test it in previously unseen environments in a zero-shot fashion. There is no prior exploration on the test set. This setting is able to clearly measure the generalizability of the navigation policy, and we evaluate our ALTR agent only under this standard testing scenario.

\subsection{Comparison with SOTA}
\begin{figure}
  \centering   
  \includegraphics[width=0.8\linewidth]{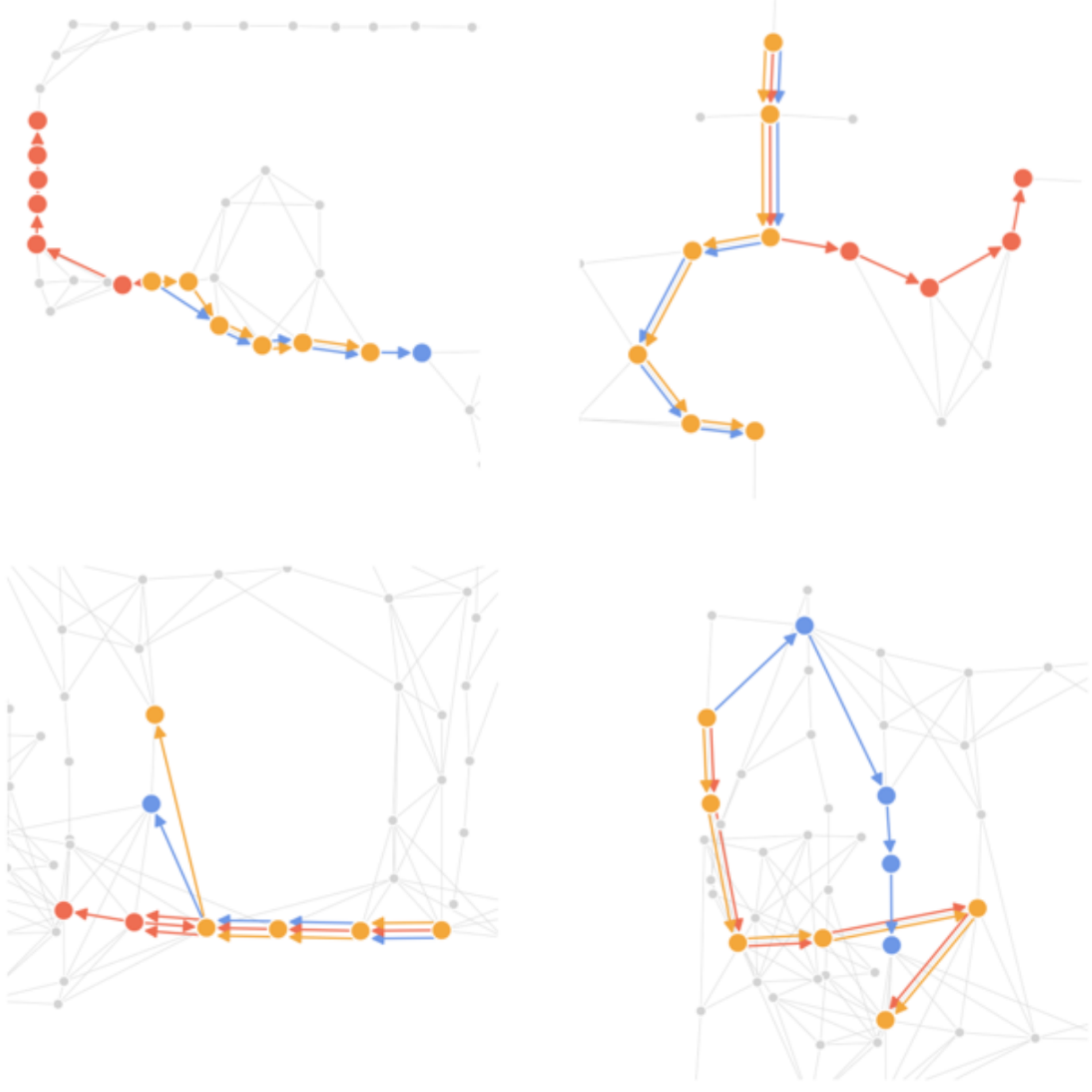}

\caption{ 
    Sample visualizations comparing reference paths (blue), paths from RCM baseline agent (red) and our ALTR agent (orange).}
    \label{fig:sample_viz}
\end{figure}

Table \ref{tab:test_submission} shows the comparison of the performance of our ALTR agent to the previous state-of-the-art (SOTA) methods on the test set of the R2R dataset, which is held out as the VLN Challenge. Our ALTR agent significantly outperforms the SOTA at the time on SPL--the primary metric for R2R--improving it by 5\% absolute measure, and it has the lowest navigation error (NE). It furthermore ties the other two best models for SR. Compared to RCM, our ALTR agent is able to learn a more efficient policy resulting in shorter trajectories to reach the goal state, as indicated by its lower path length.  Figure \ref{fig:sample_viz} compares some sample paths from the RCM baseline and our ALTR agent, illustrating that the ALTR agent often stays closer to the true path and does less doubling back compared to the RCM agent. 

\begin{table}
\setlength\tabcolsep{3.9pt}
    \centering
    \begin{tabular}{lcccc}\\[0.3em]
    Model                               & PL    & NE $\downarrow$ & SR $\uparrow$ & \underline{SPL} $\uparrow$ \\\Xhline{2\arrayrulewidth}   
    Random~\cite{Anderson:2018:VLN}      & 9.89  & 9.79           & 13.2         & 12.0  \\
    Seq-to-Seq~\cite{Anderson:2018:VLN}  & 8.13  & 7.85           & 20.4         & 18.0  \\
    \hline
    Look Before You Leap~\cite{Wang2018Look} & 9.15 & 7.53        & 25.3         & 23.0  \\
    Speaker-Follower~\cite{Fried:2018:Speaker} & 14.8 & 6.62      & 35.0         & 28.0  \\
    Self-Monitoring~\cite{Ma:2019:SelfMonitoringAgent} & 18.0 & 5.67 & \textbf{48.0}      & 35.0 \\
    Reinforced Cross-Modal~\cite{Wang:2018:RCM} & 12.0 & 6.12     & 43.1         & 38.0  \\
    The Regretful Agent~\cite{ma2019regretful} & 13.7 & 5.69 & \textbf{48.0} & 40.0 \\
    \hline
    ALTR (Ours)                                & 10.3  & \textbf{5.49}          & \textbf{48.0}        & \textbf{45.0}  \\
    \end{tabular}
    \caption{Comparison on R2R Leaderboard Test Set. Our navigation model benefits from transfer learned representations and outperforms the known SOTA on SPL. SPL and SR are reported as percentages and NE and PL in meters.}
    \label{tab:test_submission}
\end{table}

It is worth noting that the R2R leaderboard has models that use beam-search and/or explore the test environment before submission. For a fair comparison, we only compare against models that, like ours, return exactly one trajectory per sample without pre-exploring the test environment (in accordance with VLN challenge submission guidelines).

We show in the next section that our transfer learning approach improves the Speaker-Follower agent \cite{Fried:2018:Speaker}. In general, this strategy is complementary to the improvements from the other agents, so it is likely it would help others too.

\begin{table*}
\centering
\setlength\tabcolsep{4.5pt}
\begin{tabular}{lcccccccccc}\\[-0.8em]
                              &  &  & \multicolumn{4}{c}{\textbf{Validation Seen}} & \multicolumn{4}{c}{\textbf{Validation Unseen}}\\\cmidrule(lr{0.1cm}){4-7}\cmidrule(l{0.1cm}){8-11}
Method & {\taskcma} & {\tasknvs}                & PL    & NE $\downarrow$   & SR $\uparrow$   & \underline{SPL} $\uparrow$  & PL    & NE $\downarrow$   & SR $\uparrow$   & \underline{SPL} $\uparrow$   \\[0.2em]\Xhline{2\arrayrulewidth} 
Speaker-Follower \cite{Fried:2018:Speaker}                        &  -      & -      & -    & 3.36 & 66.4 & -    & -    & 6.62 & 35.5 & -\\
RCM\cite{Wang:2018:RCM}                                     & -       & -      & 12.1 & 3.25 & 67.6 & -    & 15.0 & 6.01 & 40.6 & - \\\hline

\multirow{4}{*}{Speaker-Follower (Ours)} & \xmark      &  \xmark        & 15.9 & 4.90 & 51.9 & 43.0 & 15.6 & 6.40 & 36.0 & 29.0 \\
                                        & \cmark   & \xmark      & 14.9 & 5.04 & 50.2 & 39.2 & 16.8 & 5.85 & 39.1 & 26.8 \\
                                        & \xmark   & \cmark       & 16.5    & 5.12    & 48.7    & 34.9    & 18.0    & 6.30    & 34.9    & 20.9    \\
                                        & \cmark  & \cmark & 11.3    & 4.06    & 60.8    & 55.9    & 14.6    & 6.06    & 40.0    & 31.2     \\
\hline
\multirow{4}{*}{RCM (Ours)}              & \xmark  & \xmark & 13.7 & 4.48 & 55.3 & 47.9 & 14.8 & 6.00 & 41.1 & 32.7 \\
                                        & \cmark  & \xmark & 10.2 & 5.10 & 51.8 & 49.0 & 9.5  & 5.81 & 44.8 & 42.0 \\
                                        & \xmark  & \cmark & 19.5    & 6.53    & 34.6    & 20.8    & 18.8    & 6.79    & 33.7    & 20.6   \\
                                        & \cmark  & \cmark & 13.2 & 4.68 & 55.8 & 52.7 & 9.8  & 5.61 & 46.1 & 43.0 \\
\end{tabular}
\caption{Ablations on R2R Validation Seen and Validation Unseen sets, showing results in VLN for different combinations of pre-training tasks. SPL and SR are reported as percentages and NE and PL in meters.}
\label{tab:task-ablation}
\end{table*}

\begin{table*}
\centering
\setlength\tabcolsep{4.5pt}
\begin{tabular}{cccccccccc} \\[-0.8em]
                              &  & \multicolumn{4}{c}{\textbf{Validation Seen}} & \multicolumn{4}{c}{\textbf{Validation Unseen}}\\\cmidrule(lr{0.1cm}){3-6}\cmidrule(l{0.1cm}){7-10}
Image encoder & Language encoder                    & PL    & NE $\downarrow$   & SR $\uparrow$   & \underline{SPL} $\uparrow$  & PL    & NE $\downarrow$   & SR $\uparrow$   & \underline{SPL} $\uparrow$   \\[0.2em]\Xhline{2\arrayrulewidth}
\xmark        & \xmark           & 13.7 & 4.48           & 55.3          & 47.9           & 14.8 & 6.00           & 41.1         & 32.7 \\
\cmark        & \xmark           & 15.9   & 5.05               & 50.6             & 38.2              & 14.9 & 5.94          & 42.5         & 33.1 \\
\xmark        & \cmark           & 13.8   & 4.68               & 56.3             & 46.6              & 13.5 & 5.66          & 43.9         & 35.8 \\
\cmark        & \cmark           & 13.2 & 4.68           & 55.8          & 52.7           & 9.8  & 5.61           & 46.1         & 43.0  \\

\end{tabular}
\caption{Ablations showing the effect of adapting (or not) the learned representations in each branch of our RCM agent on Validation Seen and Validation Unseen. SPL and SR are reported as percentages and NE and PL in meters.}
\label{tab:encoder-ablation}
\end{table*}

\subsection{Ablation Studies}

The first ablation study analyzes the effectiveness of each task individually in learning representations that can benefit the navigation agent. Since the agent is rewarded for reaching the goal (Eq.~\ref{eq:denseSRreward}), we expect SR results to align well with our training objective. Table \ref{tab:task-ablation} shows that agents benefit the most when initialized with representations learned on both the tasks simultaneously. When pre-trainning CMA and NVS jointly, we see a consistent 11-12\% improvement in SR for both the SF and RCM agents as well as improvement in agent's path length, thereby also improving SPL. When pre-training CMA only, we see a consistent 8-9\% improvement in SR for both the SF and RCM agents. When pre-training NVS only, we see a drop in performance. Since there are no cross-modal components to train the language encoder in NVS, training on NVS alone fails to provide a good initialization point for the downstream navigation task that requires cross-modal associations. However, pre-training with NVS and CMA jointly affords the model additional opportunities to improve visual-only pre-training (due to NVS), without compromising cross-modal alignment (due to CMA).

The second ablation analyzes the effect of transferring representations to either of the language and visual encoders. Table \ref{tab:encoder-ablation} shows the results for the RCM agent. The learned representations help the agent to generalize on previously unseen environments. When either of the encoders is warm-started, the agent outperforms the baseline success rates and SPL on validation unseen dataset. In the absence of learned representations, the agent overfits on seen environments and as a result the performance improves on the validation seen dataset. Among the agents that have at least one of the encoders warm-started, the agent with both encoders warm-started has significantly higher SPL (7\%+) on the validation unseen dataset.

The results of both the studies demonstrate that the two tasks, {\taskcma} and {\tasknvs}, learn complementary representations which benefit the navigation agent. Furthermore, the agent benefits the most when both the encoders are warm-started from the learned representations.

%% file: conclusion.tex
\section{Conclusion}
\label{sec:conclusion}

We demonstrate the model trained on two complementary auxiliary tasks, Cross-Modal Alignment ({\taskcma}) and Next Visual Scene ({\tasknvs}), learns visual and textual representations that can be transferred to navigation agents. The transferred representations improve both the SF and RCM agents in key navigation metrics. Our ALTR agent--RCM initialized with domain adapted representations--outperforms published models at the time by 5\% absolute measure. We expect our approach to be complementary to the latest state-of-the-art from Tan \etal \cite{Tan:2019:EnvironmentalDropout}.

Similar to our work, there can be other auxiliary tasks that could be designed without requiring any additional human annotations. The scoring model trained on the tasks also has additional capabilities like cross-modal alignment. We expect this could help improve methods that generate additional paired instruction-path pairs. It could also allow us to automatically segment long instruction-path sequences and thus create a curriculum of easy to hard tasks for agent training. For the future, it would be desirable to jointly train the agent with the auxiliary tasks.


%% file: main.bbl
\begin{thebibliography}{10}\itemsep=-1pt

\bibitem{Anderson:2018:Evaluation}
Peter Anderson, Angel Chang, Devendra~Singh Chaplot, Alexey Dosovitskiy,
  Saurabh Gupta, Vladlen Koltun, Jana Kosecka, Jitendra Malik, Roozbeh
  Mottaghi, Manolis Savva, and Amir~R. Zamir.
\newblock On evaluation of embodied navigation agents.
\newblock 2018.
\newblock {\tt arXiv:1807.06757 [cs.AI]}.

\bibitem{Anderson:2018:VLN}
Peter Anderson, Qi Wu, Damien Teney, Jake Bruce, Mark Johnson, Niko
  S{\"u}nderhauf, Ian Reid, Stephen Gould, and Anton van~den Hengel.
\newblock Vision-and-{L}anguage {N}avigation: Interpreting visually-grounded
  navigation instructions in real environments.
\newblock In {\em Proceedings of the IEEE Conference on Computer Vision and
  Pattern Recognition (CVPR)}, 2018.

\bibitem{Antol:2015:VQA}
S. {Antol}, A. {Agrawal}, J. {Lu}, M. {Mitchell}, D. {Batra}, C.~L. {Zitnick},
  and D. {Parikh}.
\newblock {VQA}: {V}isual {Q}uestion {A}nswering.
\newblock In {\em 2015 IEEE International Conference on Computer Vision
  (ICCV)}, pages 2425--2433, Dec 2015.

\bibitem{Bain:1999:Cloning}
Michael Bain and Claude Sammut.
\newblock A framework for behavioural cloning.
\newblock In {\em Machine Intelligence 15, Intelligent Agents [St. Catherine's
  College, Oxford, July 1995]}, pages 103--129, Oxford, UK, UK, 1999. Oxford
  University.

\bibitem{Bubic2010Prediction}
Andreja Bubić, D Cramon, and Ricarda Schubotz.
\newblock Prediction, cognition and the brain.
\newblock {\em Frontiers in human neuroscience}, 4:25, 03 2010.

\bibitem{Matterport3D}
Angel Chang, Angela Dai, Thomas Funkhouser, Maciej Halber, Matthias Niessner,
  Manolis Savva, Shuran Song, Andy Zeng, and Yinda Zhang.
\newblock {Matterport3D}: Learning from {RGB-D} data in indoor environments.
\newblock {\em International Conference on 3D Vision (3DV)}, 2017.

\bibitem{Cirik:2018:StreetView}
Volkan Cirik, Yuan Zhang, and Jason Baldridge.
\newblock Following formulaic map instructions in a street simulation
  environment.
\newblock In {\em 2018 NeurIPS Workshop on Visually Grounded Interaction and
  Language}, 2018.

\bibitem{Daftry:2016:TransferablePolicies}
Shreyansh Daftry, J.~Andrew Bagnell, and Martial Hebert.
\newblock Learning transferable policies for monocular reactive {MAV} control.
\newblock {\em CoRR}, abs/1608.00627, 2016.

\bibitem{das:2017:VisualDialog}
Abhishek Das, Satwik Kottur, Khushi Gupta, Avi Singh, Deshraj Yadav,
  Jos\'e~M.F. Moura, Devi Parikh, and Dhruv Batra.
\newblock {V}isual {D}ialog.
\newblock In {\em Proceedings of the IEEE Conference on Computer Vision and
  Pattern Recognition (CVPR)}, 2017.

\bibitem{deVries:2018:TalkWalk}
Harm de Vries, Kurt Shuster, Dhruv Batra, Devi Parikh, Jason Weston, and Douwe
  Kiela.
\newblock Talk the {W}alk: Navigating {N}ew {Y}ork {C}ity through {G}rounded
  {D}ialogue.
\newblock {\em CoRR}, abs/1807.03367, 2018.

\bibitem{Luca2016Optimal}
Massimiliano Di~Luca and Darren Rhodes.
\newblock Optimal perceived timing: Integrating sensory information with
  dynamically updated expectations.
\newblock {\em Scientific reports}, 6:28563, July 2016.

\bibitem{Donahue:2017:LTRCNCaption}
J. {Donahue}, L.~A. {Hendricks}, M. {Rohrbach}, S. {Venugopalan}, S.
  {Guadarrama}, K. {Saenko}, and T. {Darrell}.
\newblock Long-term recurrent convolutional networks for visual recognition and
  description.
\newblock {\em IEEE Transactions on Pattern Analysis and Machine Intelligence},
  39(4):677--691, April 2017.

\bibitem{Fang:2015:VisualCaptions}
H. {Fang}, S. {Gupta}, F. {Iandola}, R.~K. {Srivastava}, L. {Deng}, P.
  {Dollár}, J. {Gao}, X. {He}, M. {Mitchell}, J.~C. {Platt}, C.~L. {Zitnick},
  and G. {Zweig}.
\newblock From captions to visual concepts and back.
\newblock In {\em 2015 IEEE Conference on Computer Vision and Pattern
  Recognition (CVPR)}, pages 1473--1482, June 2015.

\bibitem{Fried:2018:Speaker}
Daniel Fried, Ronghang Hu, Volkan Cirik, Anna Rohrbach, Jacob Andreas,
  Louis-Philippe Morency, Taylor Berg-Kirkpatrick, Kate Saenko, Dan Klein, and
  Trevor Darrell.
\newblock Speaker-follower models for vision-and-language navigation.
\newblock In {\em Neural Information Processing Systems (NeurIPS)}, 2018.

\bibitem{Gillick:2018:DualEncoders}
Daniel Gillick, Alessandro Presta, and Gaurav~Singh Tomar.
\newblock End-to-end retrieval in continuous space.
\newblock 2018.
\newblock {\tt arXiv:1811.08008 [cs.IR]}.

\bibitem{Girshick:2014:Finetuning}
Ross~B. Girshick, Jeff Donahue, Trevor Darrell, and Jitendra Malik.
\newblock Rich feature hierarchies for accurate object detection and semantic
  segmentation.
\newblock In {\em 2014 {IEEE} Conference on Computer Vision and Pattern
  Recognition, {CVPR} 2014, Columbus, OH, USA, June 23-28, 2014}, pages
  580--587, 2014.

\bibitem{He:2016:Resnet}
Kaiming He, Xiangyu Zhang, Shaoqing Ren, and Jian Sun.
\newblock Deep residual learning for image recognition.
\newblock In {\em 2016 {IEEE} Conference on Computer Vision and Pattern
  Recognition, {CVPR} 2016, Las Vegas, NV, USA, June 27-30, 2016}, pages
  770--778, 2016.

\bibitem{Hemachandra:2015:VLN}
Sachithra Hemachandra, Felix Duvallet, Thomas~M. Howard, Nicholas Roy, Anthony
  Stentz, and Matthew~R. Walter.
\newblock Learning models for following natural language directions in unknown
  environments.
\newblock In {\em {IEEE} International Conference on Robotics and Automation,
  {ICRA} 2015, Seattle, WA, USA, 26-30 May, 2015}, pages 5608--5615, 2015.

\bibitem{Hermann:2019:StreeLang}
Karl~Moritz Hermann, Mateusz Malinowski, Piotr Mirowski, Andras Banki-Horvath,
  and Raia~Hadsell Keith~Anderson.
\newblock Learning to follow directions in street view.
\newblock {\em CoRR}, abs/1903.00401, 2019.

\bibitem{Hochreiter:1997:LSTM}
Sepp Hochreiter and J\"{u}rgen Schmidhuber.
\newblock Long short-term memory.
\newblock {\em Neural Comput.}, 9(8):1735--1780, Nov. 1997.

\bibitem{Hu:2016:SegmentationNLE}
Ronghang Hu, Marcus Rohrbach, and Trevor Darrell.
\newblock Segmentation from natural language expressions.
\newblock In {\em Computer Vision - {ECCV} 2016 - 14th European Conference,
  Amsterdam, The Netherlands, October 11-14, 2016, Proceedings, Part {I}},
  pages 108--124, 2016.

\bibitem{Huang:2019:Discriminator}
Haoshuo Huang, Vihan Jain, Harsh Mehta, Jason Baldridge, and Eugene Ie.
\newblock Multi-modal discriminative model for vision-and-language navigation.
\newblock In {\em Proceedings of the Combined Workshop on Spatial Language
  Understanding ({S}p{LU}) and Grounded Communication for Robotics
  ({R}obo{NLP})}, pages 40--49, Minneapolis, Minnesota, 2019. Association for
  Computational Linguistics.

\bibitem{Karpathy:2015:VisualAlignmentCaptions}
Andrej Karpathy and Fei{-}Fei Li.
\newblock Deep visual-semantic alignments for generating image descriptions.
\newblock In {\em {CVPR}}, pages 3128--3137. {IEEE} Computer Society, 2015.

\bibitem{Ma:2019:SelfMonitoringAgent}
Chih-Yao Ma, Jiasen Lu, Zuxuan Wu, Ghassan AlRegib, Zsolt Kira, Richard Socher,
  and Caiming Xiong.
\newblock Self-monitoring navigation agent via auxiliary progress estimation.
\newblock In {\em Proceedings of the International Conference on Learning
  Representations (ICLR)}, 2019.

\bibitem{ma2019regretful}
Chih-Yao Ma, Zuxuan Wu, Ghassan AlRegib, Caiming Xiong, and Zsolt Kira.
\newblock The regretful agent: Heuristic-aided navigation through progress
  estimation.
\newblock 2019.

\bibitem{Macmahon06walkthe}
Matt MacMahon, Brian Stankiewicz, and Benjamin Kuipers.
\newblock Walk the talk: Connecting language, knowledge, action in route
  instructions.
\newblock In {\em In Proc. of the Nat. Conf. on Artificial Intelligence (AAAI},
  pages 1475--1482, 2006.

\bibitem{Mao:2016:GenerationComprehensionObjects}
Junhua Mao, Jonathan Huang, Alexander Toshev, Oana Camburu, Alan~L. Yuille, and
  Kevin Murphy.
\newblock Generation and comprehension of unambiguous object descriptions.
\newblock In {\em {CVPR}}, pages 11--20. {IEEE} Computer Society, 2016.

\bibitem{matuszek:2018}
Cynthia Matuszek.
\newblock Grounded language learning: Where robotics and {NLP} meet.
\newblock In {\em Proceedings of the Twenty-Seventh International Joint
  Conference on Artificial Intelligence, {IJCAI-18}}, pages 5687--5691.
  International Joint Conferences on Artificial Intelligence Organization, 7
  2018.

\bibitem{Mirowski:2018:StreetLearn}
Piotr Mirowski, Matt Grimes, Mateusz Malinowski, Karl~Moritz Hermann, Keith
  Anderson, Denis Teplyashin, Karen Simonyan, Koray Kavukcuoglu, Andrew
  Zisserman, and Raia Hadsell.
\newblock Learning to navigate in cities without a map.
\newblock In S. Bengio, H. Wallach, H. Larochelle, K. Grauman, N. Cesa-Bianchi,
  and R. Garnett, editors, {\em Advances in Neural Information Processing
  Systems 31}, pages 2419--2430. Curran Associates, Inc., 2018.

\bibitem{Pennington:2014:GloVe}
Jeffrey Pennington, Richard Socher, and Christopher Manning.
\newblock Glo{V}e: Global vectors for word representation.
\newblock In {\em Proceedings of the 2014 Conference on Empirical Methods in
  Natural Language Processing (EMNLP)}, pages 1532--1543. Association for
  Computational Linguistics, 2014.

\bibitem{Russakovsky:2015:ImageNet}
Olga Russakovsky, Jia Deng, Hao Su, Jonathan Krause, Sanjeev Satheesh, Sean Ma,
  Zhiheng Huang, Andrej Karpathy, Aditya Khosla, Michael Bernstein,
  Alexander~C. Berg, and Li Fei-Fei.
\newblock {ImageNet Large Scale Visual Recognition Challenge}.
\newblock {\em International Journal of Computer Vision (IJCV)},
  115(3):211--252, 2015.

\bibitem{Schuster1997BidirectionalRN}
Mike Schuster and Kuldip~K. Paliwal.
\newblock Bidirectional recurrent neural networks.
\newblock {\em IEEE Trans. Signal Processing}, 45:2673--2681, 1997.

\bibitem{Serban:2018:DualEncoders}
Iulian~Vlad Serban, Ryan Lowe, Peter Henderson, Laurent Charlin, and Joelle
  Pineau.
\newblock A survey of available corpora for building data-driven dialogue
  systems: The journal version.
\newblock {\em D{\&}D}, 9(1):1--49, 2018.

\bibitem{Tan:2019:EnvironmentalDropout}
Hao Tan, Licheng Yu, and Mohit Bansal.
\newblock Learning to navigate unseen environments: Back translation with
  environmental dropout.
\newblock In {\em Proceedings of the 2019 Conference of the North American
  Chapter of the Association for Computational Linguistics: Human Language
  Technologies, {NAACL-HLT} 2019, Minneapolis, MN, USA, June 2-7, 2019, Volume
  1 (Long and Short Papers)}, pages 2610--2621, 2019.

\bibitem{Toshev:2018:SemanticNavigation}
Alexander Toshev, Arsalan Mousavian, James Davidson, Jana Kosecka, and Marek
  Fiser.
\newblock Visual representations for semantic target driven navigation.
\newblock 2018.

\bibitem{Oord2018RepresentationLW}
A{\"a}ron van~den Oord, Yazhe Li, and Oriol Vinyals.
\newblock Representation learning with contrastive predictive coding.
\newblock {\em CoRR}, abs/1807.03748, 2018.

\bibitem{Vinyals:2015:ShowAndTell}
Oriol Vinyals, Alexander Toshev, Samy Bengio, and Dumitru Erhan.
\newblock Show and tell: A neural image caption generator.
\newblock {\em 2015 IEEE Conference on Computer Vision and Pattern Recognition
  (CVPR)}, pages 3156--3164, 2015.

\bibitem{Wang:2018:VideoCV}
Xin Wang, Wenhu Chen, Jiawei Wu, Yuan-Fang Wang, and William~Yang Wang.
\newblock Video captioning via hierarchical reinforcement learning.
\newblock {\em 2018 IEEE/CVF Conference on Computer Vision and Pattern
  Recognition}, pages 4213--4222, 2018.

\bibitem{Wang:2018:RCM}
Xin Wang, Qiuyuan Huang, Asli {\c{C}}elikyilmaz, Jianfeng Gao, Dinghan Shen,
  Yuan{-}Fang Wang, William~Yang Wang, and Lei Zhang.
\newblock Reinforced cross-modal matching and self-supervised imitation
  learning for vision-language navigation.
\newblock {\em CoRR}, abs/1811.10092, 2018.

\bibitem{Wang2018Look}
Xin Wang, Wenhan Xiong, Hongmin Wang, and William~Yang Wang.
\newblock Look before you leap: Bridging model-free and model-based
  reinforcement learning for planned-ahead vision-and-language navigation.
\newblock In Vittorio Ferrari, Martial Hebert, Cristian Sminchisescu, and Yair
  Weiss, editors, {\em Computer Vision -- ECCV 2018}, pages 38--55, Cham, 2018.
  Springer International Publishing.

\bibitem{Williams:1992:PolicyGradient}
Ronald~J. Williams.
\newblock Simple statistical gradient-following algorithms for connectionist
  reinforcement learning.
\newblock {\em Machine Learning}, 8(3):229--256, May 1992.

\bibitem{2015:Xu:ShowAttendTell}
Kelvin Xu, Jimmy Ba, Ryan Kiros, Kyunghyun Cho, Aaron~C. Courville, Ruslan
  Salakhutdinov, Richard~S. Zemel, and Yoshua Bengio.
\newblock Show, attend and tell: Neural image caption generation with visual
  attention.
\newblock In {\em Proceedings of the 32nd International Conference on Machine
  Learning, {ICML} 2015, Lille, France, 6-11 July 2015}, pages 2048--2057,
  2015.

\bibitem{Yang:2016:StackedANQnA}
Zichao Yang, Xiaodong He, Jianfeng Gao, Li Deng, and Alexander~J. Smola.
\newblock Stacked attention networks for image question answering.
\newblock {\em 2016 IEEE Conference on Computer Vision and Pattern Recognition
  (CVPR)}, pages 21--29, 2016.

\bibitem{Yu:2016:VideoParagraphCaptioning}
Haonan Yu, Jiang Wang, Zhiheng Huang, Yi Yang, and Wei Xu.
\newblock Video paragraph captioning using hierarchical recurrent neural
  networks.
\newblock pages 4584--4593, 06 2016.

\bibitem{Zeiler:2014:Pretraining}
Matthew~D. Zeiler and Rob Fergus.
\newblock Visualizing and understanding convolutional networks.
\newblock In {\em Computer Vision - {ECCV} 2014 - 13th European Conference,
  Zurich, Switzerland, September 6-12, 2014, Proceedings, Part {I}}, pages
  818--833, 2014.

\bibitem{Zhu:2017:IndoorNavigation}
Yuke Zhu, Roozbeh Mottaghi, Eric Kolve, Joseph~J. Lim, Abhinav Gupta, Li
  Fei{-}Fei, and Ali Farhadi.
\newblock Target-driven visual navigation in indoor scenes using deep
  reinforcement learning.
\newblock In {\em 2017 {IEEE} International Conference on Robotics and
  Automation, {ICRA} 2017, Singapore, Singapore, May 29 - June 3, 2017}, pages
  3357--3364, 2017.

\end{thebibliography}
